%% file: top.tex
\documentclass[10pt,twocolumn,letterpaper]{article}

\usepackage{cvpr}
\usepackage{times}
\usepackage{epsfig}
\usepackage{graphicx}
\usepackage{amsmath}
\usepackage{amssymb}

\usepackage[utf8]{inputenc}
\usepackage{subfigure}
\usepackage{bbm}

\usepackage{lipsum}
\usepackage{xcolor}
\usepackage{booktabs}
\usepackage{dcolumn}
\usepackage{multirow}
\usepackage{balance}
\usepackage{enumitem}
\usepackage{float}
\usepackage{rotating}
\usepackage{caption}
\newcommand{\cev}[1]{\reflectbox{\ensuremath{\vec{\reflectbox{\ensuremath{#1}}}}}}

\usepackage[pagebackref=true,breaklinks=true,letterpaper=true,colorlinks,bookmarks=false,draft]{hyperref}

\cvprfinalcopy % *** Uncomment this line for the final submission

\def\cvprPaperID{710} % *** Enter the CVPR Paper ID here

\ifcvprfinal\fi
\begin{document}

%%%%%%%%% TITLE
\title{A Face-to-Face Neural Conversation Model}

\author{Hang Chu$^{1,2}$~~~Daiqing Li$^{1}$~~~Sanja Fidler$^{1,2}$\\
$^1$University of Toronto~~~$^2$Vector Institute\\
{\tt\small \{chuhang1122, daiqing, fidler\}@cs.toronto.edu}
}

\maketitle

\makeatletter
\def\blfootnote{\gdef\@thefnmark{}\@footnotetext}
\makeatother

%%%%%%%%% ABSTRACT
\begin{abstract}
Neural networks have recently become good at engaging in dialog. However, current approaches are based solely on verbal text, lacking the richness of a real face-to-face conversation. We propose a neural conversation model that aims to read and generate facial gestures alongside with text. This allows our model to adapt its response based on the ``mood'' of the conversation. In particular, we introduce an RNN encoder-decoder that exploits the movement of facial muscles, as well as the verbal conversation. The decoder consists of two layers, where the lower layer aims at generating the verbal response and coarse facial expressions, while the second layer fills in the subtle gestures, making the generated output more smooth and natural. We train our neural network by having it ``watch'' 250 movies.
We showcase our joint face-text model in generating more natural conversations through automatic metrics and a human study.
We demonstrate an example application with a face-to-face chatting avatar. \blfootnote{demo/data: {\color{magenta}{\url{http://www.cs.toronto.edu/face2face}}}}
\end{abstract}

%%%%%%%%% BODY TEXT
\input{intro}
\input{related}
\input{dataset}
\input{method}

\input{results}
\input{conc}

{\small
\bibliographystyle{ieee}
\bibliography{egbib}
}

\end{document}

%% file: intro.tex
\section{Introduction}
\label{sec:intro}

We make conversation everyday. We talk to our family, friends, colleagues, and sometimes we also chat with robots. Several online services employ robot agents to direct customers to the service they are looking for. Question-answering systems like Apple Siri and Amazon Alexa have also become a popular accessory. However, while most of these automatic systems feature a human voice, they are far from acting like human beings. They lack in expressivity, and are typically emotionless.

Language alone can often be ambiguous with respect to the person's mood, unless indicative sentiment words are being used. In real life, people make gestures and read other people's gestures when they communicate. Whether someone is smiling, crying, shouting, or frowning when saying ``thank you'' can indicate various feelings from gratitude to irony. People also form their response depending on such context, not only in what they say but also in how they say it. We aim at developing a more natural conversation model that jointly models text and gestures, in order to act and converse in a more natural way.

\begin{figure}[t!]
\centering
\begin{tabular}{cc}
\includegraphics[width=0.45\linewidth]{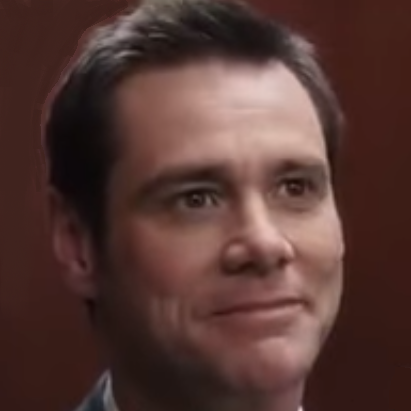} &
\includegraphics[width=0.45\linewidth]{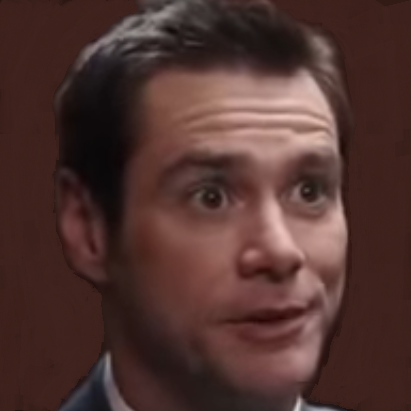}\\
\end{tabular}
\caption{\small Facial gestures convey sentiment information. Words have different meanings with different facial gestures. Saying ``\textit{Thank you}" with different gestures could either express gratitude, or irony. Therefore, a different response should be triggered.}
\end{figure}

Recently, neural networks have been shown to be good conversationalists~\cite{Vinyals15,Li15}. These typically make use of an RNN encoder which represents the history of the verbal conversation and an RNN decoder that generates a response. \cite{Li16} built on top of this idea with the aim to personalize the model by adapting the conversation to a particular user. However, all these approaches are based solely on text, lacking the richness of a real face-to-face conversation.

In this paper, we introduce a neural conversation model that reads and generates both a verbal response (text) and facial gestures. We exploit movies as a rich resource of such information. Movies show a variety of social situations with diverse emotions, reactions, and topics of conversation, making them well suited for our task. Movies are also multi-modal, allowing us to exploit both visual as well as dialogue information. However, the data itself is also extremely challenging due to many characters that appear on-screen at any given time, as well as large variance in pose, scale, and recording style.

\begin{figure*}[t!]
\centering
\hspace{-4mm}
\begin{minipage}{0.48\linewidth}
\framebox[1.0\width]{
\begin{minipage}{0.48\linewidth}
\begin{minipage}{0.33\linewidth}
\includegraphics[width=1\linewidth]{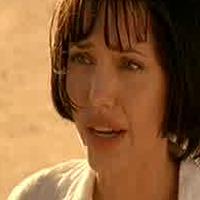}
\end{minipage}
\begin{minipage}{0.55\linewidth}
{\footnotesize I heard that already in London.}
\end{minipage}
\end{minipage}
\begin{minipage}{0.48\linewidth}
\begin{minipage}{0.33\linewidth}
\includegraphics[width=1\linewidth]{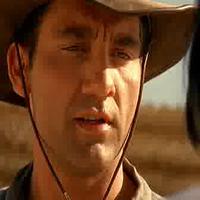}
\end{minipage}
\begin{minipage}{0.52\linewidth}
{\footnotesize You did?}
\end{minipage}
\end{minipage}}\\
\framebox[1.0\width]{
\begin{minipage}{0.48\linewidth}
\begin{minipage}{0.33\linewidth}
\includegraphics[width=1\linewidth]{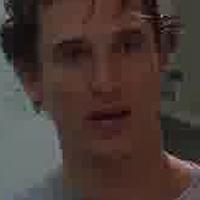}
\end{minipage}
\begin{minipage}{0.55\linewidth}
{\footnotesize Everything alright? Is this okay?}
\end{minipage}
\end{minipage}
\begin{minipage}{0.48\linewidth}
\begin{minipage}{0.33\linewidth}
\includegraphics[width=1\linewidth]{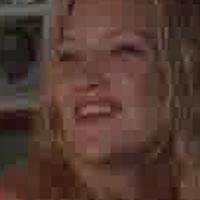}
\end{minipage}
\begin{minipage}{0.52\linewidth}
{\footnotesize Oh, no, it's more than okay.}
\end{minipage}
\end{minipage}}
\end{minipage}
\hspace{1mm}
\begin{minipage}{0.48\linewidth}
\framebox[1.0\width]{
\begin{minipage}{0.48\linewidth}
\begin{minipage}{0.33\linewidth}
\includegraphics[width=1\linewidth]{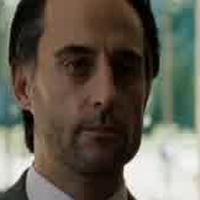}
\end{minipage}
\begin{minipage}{0.65\linewidth}
{\footnotesize I have to take the kids to the lion king, again.}
\end{minipage}
\end{minipage}
\hspace{2.5mm}
\begin{minipage}{0.48\linewidth}
\begin{minipage}{0.33\linewidth}
\includegraphics[width=1\linewidth]{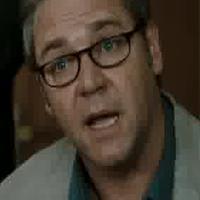}
\end{minipage}
\begin{minipage}{0.55\linewidth}
{\footnotesize Never have kids.}
\end{minipage}
\end{minipage}}\\
\framebox[1.0\width]{
\begin{minipage}{0.48\linewidth}
\begin{minipage}{0.33\linewidth}
\includegraphics[width=1\linewidth]{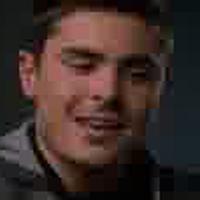}
\end{minipage}
\begin{minipage}{0.65\linewidth}
{\footnotesize I will be amazed if I can come up with something, but I will.}
\end{minipage}
\end{minipage}
\hspace{2.5mm}
\begin{minipage}{0.48\linewidth}
\begin{minipage}{0.33\linewidth}
\includegraphics[width=1\linewidth]{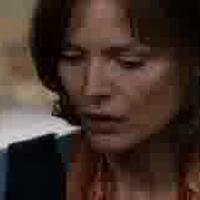}
\end{minipage}
\begin{minipage}{0.55\linewidth}
{\footnotesize I don't wanna do this any more.}
\end{minipage}
\end{minipage}}
\end{minipage}

\vspace{-2mm}
\caption{\small Example conversations from our MovieChat dataset. Each row shows two examples, left shows query face and text, right shows target face and text. Our dataset has various conversation scenarios, such as simple conversations shown in the first and second rows on the left, %common and predictable conversations shown in the third and fourth rows, as well as extremely challenging cases shown in the last row.
as well as more challenging cases shown on the right.
}
\vspace{-5mm}
\label{fig:dbexample}
\end{figure*}

%In this paper, we propose a neural conversation model that reads and generates both a verbal response (text) and facial gestures. 
Our model adopts the encoder-decoder architecture and adds gesture information in both the encoder as well as the decoder. 
%We exploit a large movie dataset~\cite{TapaswiCVPR16} to train our model.  
We exploit the FACS representation~\cite{ekman1980facial} of gestures, which allows us to effectively encode and synthesize facial gestures. Our decoder is composed of two levels, one generating the verbal response as well as coarse gesture information, and another level that fills in the details, making the generated expressions more natural.  We train our model using reinforcement learning that exploits a trained discriminator to provide the reward. We show that our model generates more appropriate responses compared to multiple strong baselines, on a large-scale movie dataset. We further showcase NeuralHank, an expression-enabled 3D chatting avatar driven by our proposed model.

\begin{figure*}[t!]
\centering
\setlength{\tabcolsep}{0cm}
\begin{tabular}{ccc}
\multirow{5}{*}{\includegraphics[height=3.5cm]{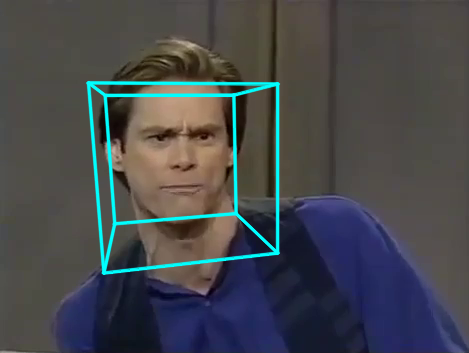}} &
\multirow{5}{*}{\includegraphics[height=3.6cm]{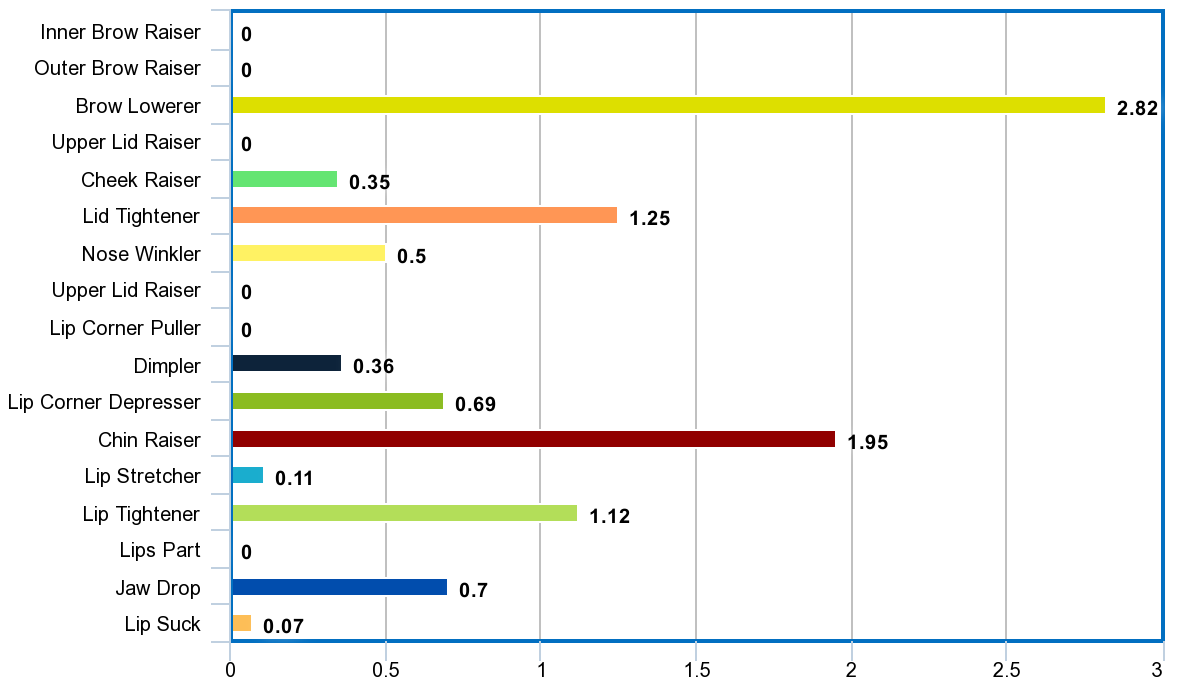}} &
\multirow{6}{*}{
\setlength{\tabcolsep}{8pt}
\setlength\extrarowheight{5pt}
\begin{minipage}{0.35\linewidth}
\vspace{1.5mm}
\begin{tabular}{l|l|l}
%\specialrule{1pt}{0pt}{2pt}
\hline
\textbf{\small query} & \textbf{\small target} & \textbf{\small \# of examples}\\
\hline
text & text & $40,200,261$\\
text+face & text & $48,475$\\
text & text+face & $48,475$\\
text+face & text+face & $24,727$\\
%\specialrule{1pt}{0pt}{8pt}
\hline
\end{tabular}
\end{minipage}
}\\
~ & ~ & ~\\
~ & ~ & ~\\
~ & ~ & ~\\
~ & ~ & ~\\
~ & ~ & ~\\
~ & ~ & ~\\
~ & ~ & ~\\
~ & ~ & ~\\[-2mm]
(a) & (b) & (c)\\
\end{tabular}
\vspace{-4mm}
\caption{\small Overview of our MovieChat database. (a) and (b) show an example frame with 3D face detection and detected FACS intensities. We obtain detections using the off-the-shelf OpenFace~\cite{baltruvsaitis2016openface} package. (c) shows the scale of our MovieChat database. Our database is by far the largest language-face conversation video dataset.}
\vspace{-3mm}
\label{fig:db}
\end{figure*}

%We make the following contributions: 
%\begin{itemize}
%\item We propose a novel conversation model that incorporates both text and gesture information. We show that our model generates more appropriate responses comparing to multiple strong baselines, on a large scale dataset.
%\item We provide a new MovieChat dataset, which is by far the largest conversation video dataset that has both language and face modalities.
%\item We showcase NeuralHank, an expression-enabled 3D chatting avatar driven by our proposed model.
%\end{itemize}

The rest of the paper is organized as follows. Sec.~\ref{sec:related} reviews the related work. In Sec.~\ref{sec:database} we introduce our dataset to facilitate face-to-face conversation modeling. In Sec.~\ref{sec:approach} we describe our approach. Sec.~\ref{sec:results} provides extensive evaluation and introduces our chit-chatting avatar. 

%% file: related.tex
\section{Related Work}
\label{sec:related}

Dialogue systems have been explored since the 60', with systems like ELIZA~\cite{eliza} and PARRY~\cite{colby} already capable of engaging in relatively complex conversations. These approaches have mainly been based on hand-coded rules, thus were not able to adapt to users and topics, and usually seemed unnatural. In~\cite{Ritter11}, the authors formulated the problem as statistical machine translation, where the goal was to ``translate'' the query posts in blogs into a response. This problem setting is typically harder than traditional translation from one language to another, since the space of possible responses is more diverse. 

Conversation modeling has recently been gaining interest due to the powerful language models learned by neural networks~\cite{Vinyals15, Li15,Li16}. \cite{Vinyals15} was the first to propose a neural conversation model, which exploited the encoder-decoder architecture. An LSTM encoder was used to represent the query sentence while the decoder LSTM generated a response, one word at a time. The model was trained on a large corpus of movie subtitles, by using each sentence as a query and the following sentence as a target during training. Qualitative results showed that meaningful responses were formed for a variety of queries. In parallel, the Skip-Thought model~\cite{skipthoughts,ZhuICCV15} adopted a similar architecture, and was demonstrated to be effective in a variety of NLP tasks as well as image-based story-telling.

Since neural conversation models typically produce short and more generic sentences, the authors in~\cite{Li15} proposed an improved objective function that encouraged diversity in the generator. In~\cite{Serban15}, the authors exploited a hierarchical encoder-decoder, where one GRU layer was used to model the history of the conversation at the sentence level, and the second level GRU was responsible for modeling each sentence at the word level. This model was extended in~\cite{Serban16} by adding latent variables aiming to capture different topics of conversation, allowing the model to achieve a higher diversity in its response. 

An interesting extension was proposed in~\cite{Li16} which aimed at personalizing conversations. The model  learned a separate embedding for each person conversationalist, jointly with dialogue. The purpose of the embedding was  to bias the decoder when generating the response. This allowed for a more natural human-like chit-chat, where the model was able to adapt to the person it was speaking to. 

Most of these works are based solely on language. However, humans often use body gestures as an additional means to convey information in a conversation. An interesting approach was proposed in~\cite{Levine09,Levine10} which aimed at synthesizing body language animations conditioned on speech using a HMM. This approach required motion capture data  recorded during several conversation sessions.

Face capture has been a long-studied problem in computer vision, with many sophisticated methods such as~\cite{baltruvsaitis2016openface,simon2017hand,hu2017avatar}. The FirstImpression dataset~\cite{biel2013youtube} was collected to facilitate the need of data in gesture recognition. Face synthesis has been widely studied in both vision and graphics communities. \cite{suwajanakorn2015makes} proposed a reconstruction algorithm that captures a person's physical appearance and persona behavior. \cite{thies2016face2face} transfers facial gesture from a source video to a target video to achieve realistic reenactment. \cite{suwajanakorn2017synthesizing} further transformed audio speech signal into a talking avatar using an RNN-based model.

In our approach, we aim to both encode and generate facial gestures jointly with language, by exploiting a large corpora of movies. Movies feature diverse conversations and interactions, and allow us to use both visual as well as dialogue information. 

%% file: dataset.tex
\section{The MovieChat Dataset}
\label{sec:database}

\begin{figure}[t!]
\centering
\includegraphics[width=0.95\linewidth]{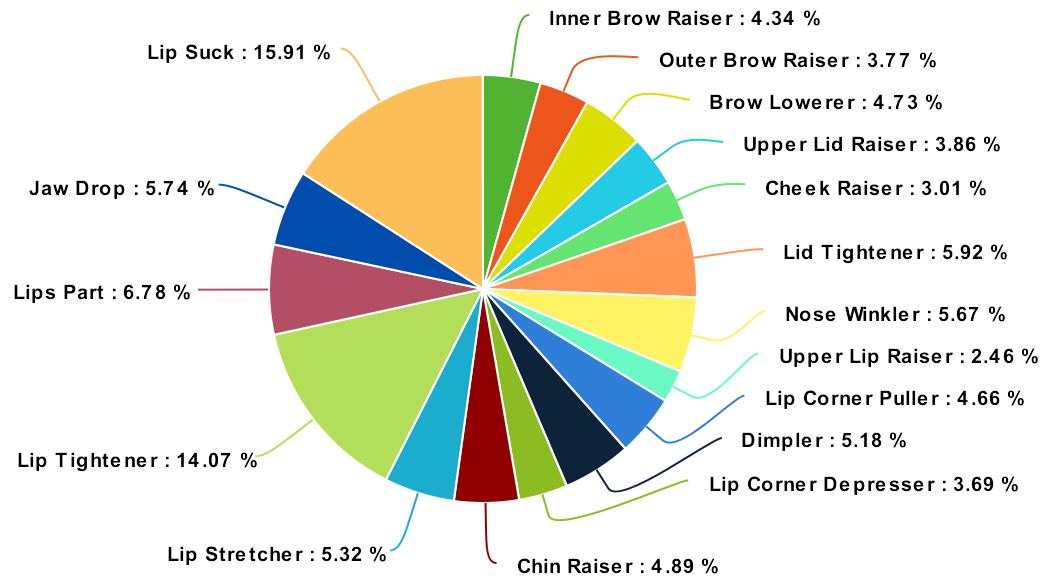}
\vspace{-4mm}
\caption{\small List of gestures recorded in the MovieChat dataset, and percentage of frames where each gesture is dominant.}
\vspace{-5mm}
\label{fig:dbstat}
\end{figure}

Datasets of considerable size are key to successfully training neural networks. In our work, we seek a dataset containing people engaging in diverse conversations, that contains both video as well as transcribed dialogues.

Towards this goal, we build the MovieChat dataset. We take advantage of the large movie collection of MovieQA~\cite{TapaswiCVPR16}, which contains clips from 250 movies, covering more than half of each movie in duration. To track 3D faces and detect facial gestures, we use the off-the-shelf OpenFace~\cite{baltruvsaitis2016openface} package. Tracking and detection runs in real-time while maintaining good accuracy. This makes processing of such a large volume of video data possible. 

However, even the best automatic face detector occasionally fails. Certain recording styles, such as the shaky and free-cam clips, make our processing more challenging. To address these problems and improve the quality of our dataset, we further divide all movies into short, single sentence clips by exploiting the time stamps stored in their subtitles file. We only keep clips where a single face is detected across all of its frames, and discard the rest of the clips. This is to avoid ambiguous dialog-face association when multiple characters appear in a single shot. Additionally, we remove fast-cut clips where the speaker's face is not fully visible throughout the clip. Finally, we also remove clips in which tracks are extremely shaky, which often suggests tracking failure. We observe significant quality improvement after these filtering steps, with only rare failure cases.

We build our final dataset with the remaining clips. We record image frames, time stamps, 3D face poses, facial gestures, and transcribed dialogues. Fig.~\ref{fig:db} shows an example, and provides statistics summarizing our dataset. Fig.~\ref{fig:dbstat} shows the recorded gestures and their statistics in our MovieChat data.

%% file: method.tex
\section{Face-to-Face Neural Conversation Model}
\label{sec:approach}
We first explain our facial gestures representation using Facial Action Coding System (FACS)~\cite{ekman1980facial}. We then describe our proposed model.

\begin{figure}[t!]%
\centering
\setlength{\tabcolsep}{4pt}
\begin{tabular}{cc}
\includegraphics[height=2.9cm]{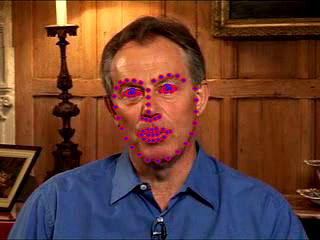} &
\includegraphics[height=2.9cm]{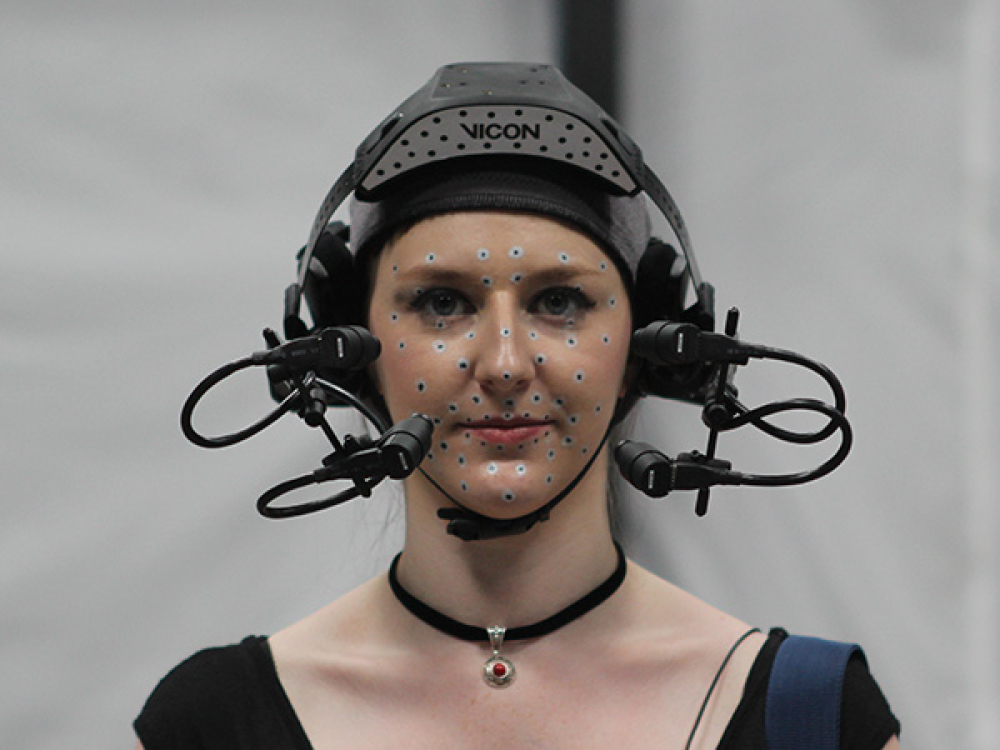}\\
\end{tabular}
\vspace{-4mm}
\caption{\small Facial Landmark (FL) systems. Left shows an example of ``in the wild'' landmarks~\cite{cootes2001active,zhu2012face,baltruvsaitis2016openface}, which fails to capture subtle gesture information. Right shows invasive motion capture landmarks~\cite{Cara}.}%
\vspace{-5mm}
\label{fig:lm}
\end{figure}

\newcommand{\mb}[1]{\mathbf{#1}}
\newcommand{\mt}[1]{\mathcal{#1}}
\newcommand{\mc}[1]{\mathcal{#1}}

\begin{figure*}[t!]%
\centering
\includegraphics[width=0.94\linewidth]{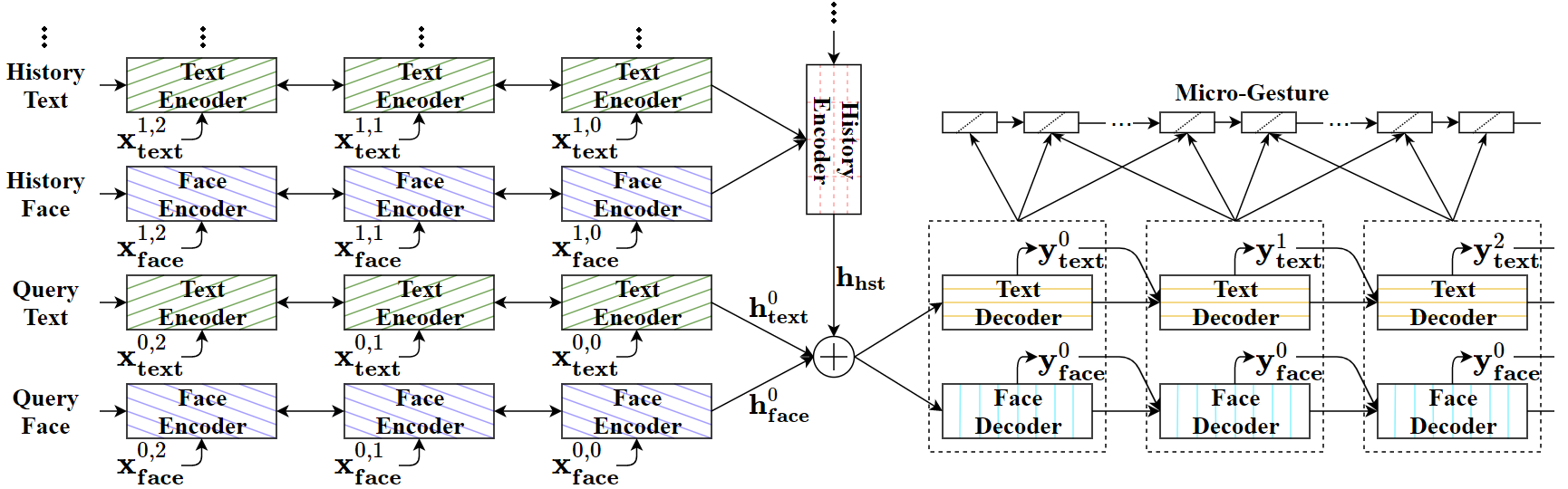}
\caption{\small Our face-to-face conversation model. Our model consists of $6$ RNNs shown in different colors. First, the text-face sequences of query and conversation history are encoded by text and face encoders (only one history sentence is depicted). History sentence encodings are further encoded by the history encoder. Next, encodings are added to form the context vector, which the text and face decoders are conditioned on. Finally, we generate frame-level, micro-gesture animation controls based on the word decodings.}%
\vspace{-0mm}
\label{fig:overview}
\end{figure*}

\subsection{FACS Gesture Representation}
Various approaches are available for representing gesture numerically, e.g. Six Universal Expressions (SUE)~\cite{black1997recognizing}, Facial Landmarks (FL)~\cite{cootes2001active,zhu2012face}, and FACS~\cite{ekman1980facial}. 

SUE~\cite{black1997recognizing} categorizes gesture into six emotions: anger, disgust, fear, happiness, sadness and surprise. It is effective in encoding high-level emotion, but it is overly abstract to describe detailed gestures. Each emotion involves a combination of up to $6$ muscle movements, making it difficult for face synthesis and animation.

FL~\cite{cootes2001active,zhu2012face} represents gesture using landmark points. Typically, $68$ points are used to track corner-edge keypoint positions of the face. Compared to SUE, FL carries more details. However, FL has two disadvantanges. First, FL does not contain complete gesture information. The cheek and forehead regions, which are texture-less but contain many muscles, are usually missing. Second, FL is anatomically redundant. $5$ landmarks are used to outline one brow, while its underlying motion is lower dimensional that involves $2$ muscle intensity values. Therefore, FL is less desirable for our task. It should be noted there are variations of FL that places landmarks across all muscles evenly. They are widely used in motion capture, e.g. Cara~\cite{Cara} in Figure~\ref{fig:lm}. This FL system requires visible marks on the character face, thus making large scale data collection difficult. 

We adopt FACS~\cite{ekman1980facial} in this paper. Particularly, we use $18$ action unit each controls a face muscle, as well as $3$ dimensions to represent the 3D head pose. Compared to the SUE and FL, FACS not only captures subtle detail gestures, but also produces highly interpretable gesture representation which makes animation simple and straight-forward. We detect FACS from images using the off-the-shelf OpenFace software~\cite{baltruvsaitis2016openface}.

\subsection{Face-to-Face Conversation Model}
Following previous work on conversation modeling~\cite{Serban16,sutskever2014sequence}, we adopt the RNN encoder-decoder architecture, but adapt it to our face-to-face conversation task. Our proposed model consists of $6$ RNN modules that capture and generate information across different modalities and resolutions. Fig.~\ref{fig:overview} provides an overview of the model. Overall, our model is an encoder-decoder framework that is trained with RL and GAN.

{\bf Notation.} Our algorithm takes a series of paired text and gesture sequences as input. These represent the query sequence as well as a recent conversation history of $N$ sequences. Here, let ${\bf x}^0$ denote the current query sequence, while ${\bf x}^n$ indexes the $n$-th sequence of the current history. We will use subscripts $\mb{text}$ and $\mb{face}$ to denote the data from the two modalities. 
%We will use $\mb^{n}$
%Let $\{L_0, L_1, ... L_N\}$ denote the number of words in each sentence, where $L_0$ is the word length of query (source), $L_1$ and $L_N$ are word lengths for the most recent and earliest sequences in the history. The reverse-order $\ell$-th word in the $n$-th history sentence ($n$=0 for query sentence) is therefore a pair of $(\mb{x}_{\mb{text}}^{L_{n}-\ell},\mb{x}_{\mb{face}}^{L_{n}-\ell})$, where $n\in [0,N]$ and $\ell\in [0,L_n]$. Similarly, we denote the ground truth answer (target) sequence as $(\hat{\mb{y}}_{\mb{text}}^{L_{*}-\ell},\hat{\mb{y}}_{\mb{face}}^{L_{*}-\ell})$, where $\ell\in [0,L_{*}]$, $L_{*}$ being the answer word length.

{\bf Sentence encoders.} We synchronize text and gestures at the word level. 
%Let ${\bf x}^{n}$ represent a one-hot encoding of the 
%$\ell$-th word in sentence ${\bf s}_\mb{text}^n$. 
Let $\mb{x}_{\mb{text}}^{n,\ell}$ represent a one-hot encoding of the $\ell$-th word in the $n$-th sentence.  
To keep the representation consistent and to simplify the multi-dimensional gesture data, we set $\mb{x}_{\mb{face}}^{{n},\ell}$ as a similar one-hot encoding of the closest gesture template. Gesture templates are obtained via k-means ($k=200$) clustering of all gestures in the training set. We define our sentence-level encoders as bidirectional RNNs, i.e.
\begin{equation}
\begin{aligned}
\mb{h}_{\mb{text}}^{{n}} &= \textrm{BiLSTM}(\{\mb{x}_{\mb{text}}^{{n},\ell}\}_\ell)\\
\mb{h}_{\mb{face}}^{{n}} &= \textrm{BiLSTM}(\{\mb{x}_{\mb{face}}^{{n},\ell}\}_\ell)
\end{aligned}
\end{equation}
%where $\mb{h}^{}$ denotes the hidden state at time step $\ell$, and $\mb{F}$ denotes the RNN cell function.
where the BiLSTM computes the forward and backward sentence encodings $\vec{h}^n$ and $\cev{h}^n$, respectively, concatenates them, and applies a linear layer on top. 

{\bf History encoder.} To model the context of the conversation, we take a history of $N$ sequences (excluding query), and add another bidirectional RNN over the encoded text and gesture sequences:
\begin{equation}
\begin{aligned}
\mb{h}_{\mb{hst}} = \textrm{BiLSTM}(\{\mb{h}_{\mb{text}}^{{n}}\oplus \mb{h}_{\mb{face}}^{{n}}\}_n)
\end{aligned}
\end{equation}
where  $\oplus$ denotes vector concatenation.

{\bf Sentence decoders.} We use two decoders that generate the target text and gesture sequences. The output follows the same one-hot encoding as query and history sentences. We condition the target sentence generation on the joint text-face-history context vector, which is obtained by the summation of the query text encoding, query face encoding, and history encoding. Concretely,
\begin{equation}
\begin{aligned}
\mb{h}_{\mb{enc}} = \mb{h}_{\mb{text}}^{{0}}+\mb{h}_{\mb{face}}^{{0}}+\mb{h}_{\mb{hst}}
\end{aligned}
\end{equation}
where $\mb{h}_{\mb{enc}}$ is the final encoding that we condition our generation decoders on. We use two independent single-directional RNN decoders, i.e.
\begin{equation}
\begin{aligned}
\mb{h}_{\mb{dec}}^{\ell} &= \textrm{LSTM}(\mb{h}_{\mb{dec}}^{\ell-1}\mid \mb{h}_{\mb{enc}},\mb{y}^{\ell-1})\\
\mb{y}^{\ell} &= \underset{\mb{y}}{\rm{argmax}} \ p(\mb{y}\mid \mb{h}_{\mb{dec}}^{\ell}),
\end{aligned}
\end{equation}
where $\mb{y}^{\ell}$ is either $\ell$-th output word or gesture, and $p$ computes a softmax over a linear layer on top of the hidden state.
%where we pad the beginning with the \texttt{\small begin\_of\_sent} token and terminates the decoding at the \texttt{\small end\_of\_sent} token. The face decoder follows the same definition.

\newcommand{\fir}[1]{\textbf{\textcolor{red}{#1}}}
\newcommand{\sece}[1]{\textbf{\textcolor{blue}{#1}}}

\begin{table*}[t!]
\centering
\setlength{\tabcolsep}{4pt}
\setlength{\extrarowheight}{-2pt}
{\small
{\tt
\begin{tabular}{ll|lll|lll|lll}
%\specialrule{1pt}{0pt}{2pt}
~ & ~ & \multicolumn{3}{c|}{{\rm beam=1}} & \multicolumn{3}{c|}{{\rm beam=3}} & \multicolumn{3}{c}{{\rm beam=5}}\\
~ & {\rm \textit{perp.}} & {\rm \textit{pre. \%}} & {\rm \textit{rec. \%}} & {\rm $F1$} & {\rm \textit{pre. \%}} & {\rm \textit{rec. \%}} & {\rm $F1$} & {\rm \textit{pre. \%}} & {\rm \textit{rec. \%}} & {\rm $F1$}\\
\hline
{\rm Text~\cite{skipthoughts,sutskever2014sequence}} & 32.53 & 23.18 & 15.58 & 17.12 & \fir{25.00} & 17.13 & 18.62 & \sece{24.70} & 16.91 & 18.34\\
{\rm Text+RandFace} & 32.65 & 22.92 & 15.99 & 17.27 & 24.74 & 17.32 & 18.57 & \fir{24.71} & 17.82 & 18.84\\
{\rm Text+Face} & \sece{30.17} & 24.25 & 17.52 & 18.69 & \sece{24.78} & 18.60 & 19.40 & 24.34 & 18.74 & 19.37\\
\hline
{\rm History-RNN~\cite{serban2015building}} & 31.15 & 23.99 & 19.46 & 19.59 & 23.79 & 20.11 & 19.67 & 23.37 & \fir{20.50} & 19.68\\
{\rm History-FC} & 30.39 & 24.49 & 19.61 & 19.88 & 24.38 & \underline{\fir{20.50}} & \fir{20.14} & 23.70 & 20.45 & 19.91\\
\hline
{\rm Ours-MLE} & \underline{\fir{30.08}} & 25.16 & 19.72 & 20.17 & 24.50 & \sece{20.32} & \sece{20.11} & 23.75 & \sece{20.47} & 19.89\\
{\rm Ours-F1} & 31.91 & \sece{25.16} & \sece{20.24} & \sece{20.42} & 24.48 & 20.26 & 20.02 & 24.06 & 20.33 & \sece{19.96}\\
{\rm Ours-GAN} & 31.60 & \underline{\fir{25.23}} & \fir{20.19} & \underline{\fir{20.44}} & 24.56 & 20.31 & 20.08 & 24.11 & 20.38 & \fir{19.97}\\
%\specialrule{1pt}{2pt}{8pt}
\end{tabular}
}
}
\vspace{-3mm}
\caption{\small The mind-reading text results on text. Second column lists word \textit{perplexity} (lower the better). Third to last columns list unigram \textit{precision}, \textit{recall}, and $F1$-\textit{score} (higher the better) across different beam search size. For each column, we mark the \fir{best} and \sece{second best} results in red and blue color. We underscore the \underline{\fir{overall best}} result across all methods and all beam sizes.}
\vspace{-2mm}
\label{tab:face4lan}
\end{table*}

\begin{table*}[t!]
\centering
\setlength{\tabcolsep}{4pt}
\setlength{\extrarowheight}{-2pt}
{\small
{\tt
\begin{tabular}{ll|lll|lll|lll}
%\specialrule{1pt}{0pt}{2pt}
~ & ~ & \multicolumn{3}{c|}{{\rm beam=1}} & \multicolumn{3}{c|}{{\rm beam=3}} & \multicolumn{3}{c}{{\rm beam=5}}\\
~ & {\rm \textit{perp.}} & {\rm \textit{pre. \%}} & {\rm \textit{rec. \%}} & {\rm $F1$} & {\rm \textit{pre. \%}} & {\rm \textit{rec. \%}} & {\rm $F1$} & {\rm \textit{pre. \%}} & {\rm \textit{rec. \%}} & {\rm $F1$}\\
\hline
{\rm Face~\cite{sutskever2014sequence}} & 18.98 & 26.48 & 9.83 & 12.96 & 22.41 & 8.18 & 10.82 & 20.74 & 7.55 & 10.02\\
{\rm Face+RandText} & 18.94 & 26.63 & 10.01 & 13.15 & 22.54 & 8.15 & 10.82 & 20.20 & 7.43 & 9.80\\
{\rm Face+Text} & 17.20 & 29.46 & 10.89 & 14.41 & 25.84 & 9.46 & 12.57 & 24.82 & 9.14 & 12.15\\
\hline
{\rm History-RNN~\cite{serban2015building}} & 20.30 & 20.84 & 7.33 & 9.81 & 20.84 & 7.33 & 9.81 & 20.84 & 7.33 & 9.81\\
{\rm History-FC} & 20.26 & 20.86 & 7.35 & 9.83 & 20.81 & 7.33 & 9.80 & 20.84 & 7.33 & 9.81\\
\hline
{\rm Ours-MLE} & \underline{\fir{17.18}} & 35.81 & 13.74 & 18.07 & \fir{30.44} & \fir{11.43} & \fir{15.10} & \sece{28.25} & \sece{10.58} & 13.49\\
{\rm Ours-F1} & 17.20 & \underline{\fir{36.17}} & \underline{\fir{13.92}} & \underline{\fir{18.28}} & 30.42 & \sece{11.43} & \sece{15.09} & \fir{28.30} & \fir{10.63} & \fir{14.06}\\
{\rm Ours-GAN} & \sece{17.19} & \sece{36.06} & \sece{13.85} & \sece{18.20} & \sece{30.43} & 11.38 & 15.05 & 28.12 & 10.52 & \sece{13.92}\\
%\specialrule{1pt}{2pt}{8pt}
\end{tabular}
}
}
\vspace{-3mm}
\caption{\small The mind-reading test results on gesture. Legend same as Table \ref{tab:face4lan}.}
\vspace{-0mm}
\label{tab:lan4face}
\end{table*}

{\bf Micro-gesture generator.} The output of our gesture decoder is a gesture template (cluster). We observe that although templates are sufficient for representing semantics, they are insufficient for synthesizing vivid, high framerate animations. We use the micro-gesture module to fill in this resolution gap, effectively interpolating between the consecutive discrete gestures, and adding relevant variations to the final gestures. We define this module as a frame-level RNN. For the $t$-th frame, we synthesize its micro-gesture based on the two most adjacent words, i.e.
\begin{equation}
\begin{aligned}
%\mb{h}_{\mb{micro}}^{t} = \textrm{LSTM}(\mb{y}_{\mb{text}}^{t}\oplus \mb{y}_{\mb{face}}^{t}\mid \mb{h}_{\mb{micro}}^{t-1})
\mb{h}_{\mb{micro}}^{t} = \textrm{LSTM}( \mb{h}_{\mb{micro}}^{t-1}\mid \mb{y}_{\mb{text}}^{t-\delta}\oplus \mb{y}_{\mb{face}}^{t-\delta},\mb{y}_{\mb{text}}^{t+\delta}\oplus \mb{y}_{\mb{face}}^{t+\delta})
\end{aligned}
\end{equation}
where $\delta$ denotes the interpolation interval ($t-\delta$ indexes the previous word, and $t+\delta$ the next one). We obtain the frame-level gesture by linearly regressing $\mb{h}_{\mb{micro}}^{t}$ to each individual gesture dimension. These gesture values directly control the muscle intensities that drive a 3D avatar.

{\bf Policy gradient optimization.} As typical, we train the model with the cross-entropy loss. However, the default cross-entropy training suffers from exposure bias, as the model is only exposed to ground truth samples during training. For our decoder networks, we alleviate this problem by optimizing directly for the desired metrics using policy gradient optimization. In this setting, the \textit{policy} takes the form of a decoder RNN, and an \textit{action} is a sentence sampled from the policy denoted by $\tilde{\mb{y}}$ (for either $\mb{text}$ or $\mb{face}$). Our goal is to expose the model to more samples of $\tilde{\mb{y}}$, and discover a policy to achieve higher \textit{reward} under the metric of choice evaluated at the end of the sequence (e.g. $F1$-score). This is denoted by $\mb{R}(\tilde{\mb{y}},\hat{\mb{y}}_{\mb{gt}})$, where $\hat{\mb{y}}_{\mb{gt}}$ is the ground truth sequence. %Denoting the sequnce generator as $\mb{F}^{\theta}$ with $\theta$ denoting its parameters. 
The policy gradient (using a single sample) is computed as follows ($\mb{h}$ short for $\mb{h}_{\mb{enc}}$): 
\begin{equation}
\begin{aligned}
%\nabla J_{\mb{text}} = [\mb{R}(\tilde{\mb{y}}_{\mb{text}}| \mb{h}_{\mb{enc}})-\mb{b}]\nabla \log \mb{F}_{\mb{text\_dec}}(\tilde{\mb{y}}_{\mb{text}}| \mb{h}_{\mb{enc}})
\nabla J_{\mb{pg}}(\theta) = [\mb{R}(\tilde{\mb{y}},\hat{\mb{y}}_{\mb{gt}})-\mb{b}]\nabla_\theta \log p_{\theta}(\tilde{\mb{y}}| \mb{h})
\end{aligned}
\end{equation}
where $J_{\mb{pg}}$ is the objective function, and $\mb{b}$ is the baseline to help reduce the variance of the gradients. 
We follow~\cite{rennie2016self}, and compute the baseline by greedy decoding  conditioned on the same $\mb{h}$. Computing the baseline in this way mimics the inference strategy at test time, thus obtaining positive gradients whenever the sampled sequence scores higher than the current greedy sequence. 

For computing the reward, we will exploit standard metrics such as $F1$-score, as well as learned reward functions as explained next.
%, and $\tilde{\mb{y}}_{\mb{text}}$ is a sample from the decoder.

{\bf Reward via an adversarial discriminator.} Conversation models suffers from dull responses~\cite{li2016deep}, while diverse dialogues are preferred in practical scenarios. We address this problem by using a sequence GAN~\cite{yu2017seqgan}, following the idea of~\cite{dai2017towards} for captioning. The \textit{generator} is our decoder network, while the \textit{discriminator} is another network that distinguishes whether the resulting sequence is machine-generated (fake) or real. We can formulate this as a minmax problem, i.e.
\begin{equation}
\begin{aligned}
\underset{\theta}{\rm{min}}\  \underset{\eta}{\rm{max}}\  J_{\mb{gan}}(p_{\theta},\mb{D}_{\eta})
\end{aligned}
\end{equation}
where $\mb{D}$ is the discriminator producing probability value $[0,1]$, $\eta$ being its parameters. Specifically, the GAN objective is defined as
\begin{equation}
\begin{aligned}
%\mathbb{E}_{\mb{y}\sim\hat{\mb{y}}_{\mb{h}}}[\rm{log}\mb{D}^{\eta}(\mb{h}_{\mb{enc}},\mb{y})]+\mathbb{E}[\rm{log}(1-\mb{D}^{\eta}(\mb{h}_{\mb{enc}},\mb{F}^{\theta}(\mb{h})))]
%\mathbb{E}_{\mb{y}\sim\hat{\mb{y}}_{\mb{h}}}[\rm{log}\mb{D}^{\eta}(\mb{h},\mb{y})]+\mathbb{E}[\rm{log}(1-\mb{D}^{\eta}(\mb{h},\mb{F}^{\theta}(\mb{h})))]
J_{\mb{gan}} = \mathbb{E}_{\hat{\mb{y}}_{\mb{gt}}}[\rm{log}\,\mb{D}_{\eta}(\mb{h},\hat{\mb{y}}_{\mb{gt}})]+\mathbb{E}_{\tilde{\mb{y}}\sim p_\theta}[\rm{log}\,(1-\mb{D}_{\eta}(\mb{h},\tilde{\mb{y}}))]
\end{aligned}
\end{equation}
The discriminator is conditioned on $\mb{h}$, trying to both learn what good sequences are and their consistency with the query sequences. 
When training the discriminator we follow~\cite{dai2017towards}, to also add mis-matched query-sequence pairs in the discriminator training step to improve the generation's semantic relevance. We use $\mb{D}$ to compute the reward for policy-gradient optimization.
%This can be viewed as an extended version of policy gradient, where the reward function is replaced by a discriminator that is simultaneously trained.

{\bf Implementation details.} We substantiate all LSTMs with a 1024-d LSTM cell~\cite{hochreiter1997long} on top of a 512-d embedding layer, followed by a linear layer with hyperbolic tangent non-linearity to compute the final encoding. Our GAN discriminator is implemented as a 3-layer, 512-d MLP that takes sentence encoding and context vectors as inputs.

Nested hierarchical neural networks are difficult to train from scratch in an end-to-end fashion. We observe the same for our model. To train our model successfully, we first pre-train our text and face encoders on single sequence corpora. Then we freeze the encoder modules and use them to generate setence-level encodings, which is used to pre-train our history model. Similarly, we pre-train decoders to make them familiar with the context. After all modules are pre-trained, we jointly finetune the entire network.

To train our decoder, we adopt the MIXER~\cite{ranzato2015sequence} strategy. We initialize the policy network via MLE. Then we gradually anneal MLE steps and blend in RL steps temporally. We keep this process until all time steps are replaced by RL. To train our discriminator, we mix the same ratio of sampled sequences, ground truth sequences, and mis-matched ground truth. In PG training, we observe that balanced positive and negative rewards are also helpful for the training process. In our case, we randomly discard samples until average reward is equal to the baseline reward. We use clipped gradient descent in our pre-training steps, and Adam~\cite{kingma2014adam} in all other training steps.

We pre-train our micro-gesture module on the FirstImpression dataset~\cite{biel2013youtube}, which contains close-up talking videos that allows high precision tracking of micro-gesture. To synchronize words and gestures at the frame level, we perform speech recognition with Bluemix, and force the alignment with existing transcripts with the Smith-Waterman algorithm~\cite{smith1981identification}. We reduce the jittering effect of our final generation using an online Savitzky-Golay filter~\cite{savitzky1964smoothing}.

%% file: results.tex
\section{Experiments}
\label{sec:results}
%We design two experiments to evaluate our model, from both quantitative and qualitative perspectives. In our experiments, we divide our dataset into train, validation, and test sets with ratio 4:1:1. We tune model hyper-parameters and perform early stopping by referring to loss on the validation set.

We evaluate our model through automatic metrics with a ``mind-reading'' test, and through a human study with a NeuralHank chatting avatar controlled by our model. We randomly split MovieChat into 4:1:1 train-val-test, and keep the split in all experiments.  

\input{qualitative}

\subsection{The Mind-Reading Test}
In the first experiment, we evaluate how well the model's generation matches with the ground truth target text and gesture sequences. This reflects the model's ability to produce appropriate, human-like responses. We note that this is an extremely challenging task, particularly for producing fair evaluation. Due to the multi-modal nature of chit-chat conversations, there exists many plausible responses to the same query, and the ground truth only represents one mode among many. Therefore, we refer to this evaluation as the mind-reading test.

We evaluate the model at both word and sentence level. At the word level, we evaluate the \textit{perplexity}, i.e. the likelihood of generating the correct next target word, given the source and correct previous words in the target sequence. This measures coherence of the textual and facial language models. At the sentence level, we evaluate  \textit{precision}, \textit{recall}, and $F1$-\textit{score} between the words in generated sentences and ground truth. %While our model also outperforms the baseline in the more standard BLEU metric, we feel that due to the hardness of the task, evaluating the keyword correctness  is a slightly better measure. 
%Despite our method also achieves higher BLEU than baselines, we instead choose these metrics because we rarely observe grammar mistake in generation due to extensive language model pre-training, and the proportion of times certain keywords are correctly guessed becomes a better reflection of mind-reading capability.
%Note that this metric is somewhat similar to BLEU-1

We compare our approach to five baselines: {\bf 1.)} Text(Face): The classic Seq2Seq~\cite{skipthoughts,sutskever2014sequence} method that uses single modality only (either text or face) and only the query sequence (no history); {\bf 2.)} Text+Face(Face+Text): Two encoders for both text and face query senquences without history; {\bf 3.)} Text+RandFace(Face+RandText): Same model as previous but trained with randomized face(text) query sentences; {\bf 4.)} History-RNN: Modeling conversation history as well as query text(face) using a hierarchical RNN, which is similar to to~\cite{serban2015building}; {\bf 5.)} History-FC: Same as previous but directly connects history sentences to the decoder with fully connected layers. This exploits the potential in conversation history, at the cost of inflexibility to history length $N$ and significantly heavier models. For our model, we compare Ours-MLE, Ours-F1 ($F1$-score as reward), and Ours-GAN. We use beam search with varying sizes for all methods.

From Table~\ref{tab:face4lan}, \ref{tab:lan4face}, it can be seen that our method achieves the best performance. Due to non-overlapping conversation scenes between data splits, the improvement of our methods is meaningful and generalizable. Therefore, our experiments quantitatively prove the common intuition that seeing the face makes understanding the conversation easier and better, backing up the main argument of this paper. Our base model can be further improved using reinforcement and adversarial training. GANs do not achieve better automatic metric score than directly setting metric reward for PG. However, GANs are able to generate more diverse and interesting responses, as we will later show in Sec.~\ref{sec:hank}. This finding is in accordance with image captioning~\cite{dai2017towards}.

{\bf The role of gestures}. In Table~\ref{tab:face4lan}, Text+Face outperforms only Text, indicating that gesture information helps text understanding. Text+RandFace does not achieve significant improvement despite a slightly better $F1$-score. This verifies the improvement of Text+Face is indeed due to the effectiveness of gesture data, instead of the additional encoder stream. This justifies our argument that gesture information is useful for text understanding. Our method outperforms both History-RNN and History-FC, showing the compatibility between gesture and history information.

{\bf The role of text}. Similarly, from Table~\ref{tab:lan4face} it can be seen that in understanding and generating face gestures, text information is helpful. This confirms the mutual benefit between both text and gesture information.

{\bf The role of history}. In both text and face, History-FC outperforms History-RNN. This indicates that there is still room for further improvement for better history encoders. However, History-RNN remains the preferable option, for its smaller model size and flexibility to varying history length, which is important in practical scenarios. Interestingly, history method outperforms Text+Face in text mind reading, indicating that multiple sentences of text history is more helpful than a query face sequence. It is the opposite in gesture mind reading, indicating that when guessing facial gestures, seeing the source face and react accordingly can be more helpful than knowing a series of text-only history sentences. This can be also partially due to the nature of movie data, where both source and target can be conveyed by the same character.

\input{hankfig}

\subsection{The NeuralHank Chatbot}\label{sec:hank}
Here, we test how our model's performance in the eyes of real human users. We create a virtual chatbot named NeuralHank, that is controlled by our model. This experiment  aims to demonstrate the more realistic potential of our model and provides a pilot study towards new applications, e.g. AI assistants and gaming/HCI.  

In NeuralHank, we ensemble a series of off-the-shelf packages to convert our model's generation into a real talking avatar. We use Microsoft Speech API to render text as audio, while also keeping record of viseme time tags. We then render FACS gestures using Maya's Facial Animation Toolset, with its default character Hank. For Hank's lip motion, we simply use the viseme event record with tangent interpolation over time. In training, we continue for a few more epochs after the early stopping point until training loss is below a certain threshold. We found this makes the model's generation more particular, which is helpful for building a lively avatar.

We compare three methods: {\bf 1.)} \textit{noMicro-noGAN} that uses beam search without GANs, and only word-level face decoder without micro-gesture RNN; {\bf 2.)} \textit{Micro-noGAN} that uses the micro-gesture RNN, and sampling without GANs; {\bf 3.)} \textit{Ours} as our full model.

We conduct a human study via Amazon Mechanical Turk, by asking participants to rate different methods' responses on the same query. We ask participants to choose the more interesting and natural response, in terms of text, gestures, and overall. For query subjectivity, we randomly choose $65$ query sentences from our held-out test set and run all methods using them as inputs. Most participants are not well-trained experts. It is important to make our study easy to follow. To achieve this, we only display a pair of different methods' generations in random order, instead of showing all three together. We also intentionally set query text as the most important information, and set query gesture and history as zero. This makes our task easy to understand, while not affecting the fairness of comparison because the methods only differ on the decoder side.

\begin{table}[t!]
\centering
{\small
{\tt
\begin{tabular}{l|lll|l}
~ & {\rm \textit{text \%}} & {\rm \textit{face \%}} & {\rm \textit{overall \%}} & ~\\
\hline
{\rm \textit{noMicro-noGAN}} & 48.8 & 39.0 & 46.2 & \multirow{6}{*}{\rm \textit{pairwise}}\\
{\rm \textit{Micro-noGAN}} & \textbf{51.2} & \textbf{61.0} & \textbf{53.8} &\\
\cline{1-4}
{\rm \textit{noMicro-noGAN}} & 44.8 & 35.3 & 42.4 &\\
{\rm \textit{Ours}} & \textbf{55.2} & \textbf{64.7} & \textbf{57.6} &\\
\cline{1-4}
{\rm \textit{Micro-noGAN}} & 46.1 & 48.8 & 46.7 &\\
{\rm \textit{Ours}} & \textbf{53.9} & \textbf{51.2} & \textbf{53.3} &\\
\hline
{\rm \textit{noMicro-noGAN}} & 31.5 & 25.0 & 29.8 & \multirow{3}{*}{\rm \textit{accumu.}}\\
{\rm \textit{Micro-noGAN}} & 32.5 & 36.8 & 33.5 &\\
{\rm \textit{Ours}} & \textbf{36.0} & \textbf{38.2} & \textbf{36.6} &\\
\end{tabular}
}
}
\vspace{-3mm}
\caption{\small AMT user study on interestingness and naturalness. The evaluation is conducted in form of pairwise comparison. We further accumulate number of votes for different methods.}
\vspace{-2mm}
\label{tab:amt}
\end{table}

We request 10 Turkers for each sample. This results in 5850 answers from 37 unique participants. %Due to the novelty of our study and the rush tendency of users, our evaluation concept is not well understood by all participants. To address this, we stress our goal of identifying \textit{interesting} and \textit{natural} responses in task description. 
We further use exam questions to filter out the noisy participant responses. The questions are verified samples where one answer is obviously better, e.g. a spot on, grammar error free, and fun sentence, versus a simple and boring yes/no answer. %Table~\ref{tab:amt} lists the result.

It can be seen from Table~\ref{tab:amt} that micro-gesture significantly improves gesture quality. Our full model with adversarial training achieves the best user rating from all three perspectives. Compared to no-GAN methods that tend to produce universally correct but less interesting responses, GAN methods produces generally more diverse and interesting responses. However, GAN methods also suffer from occasional confusing or offensive responses. 

Fig.~\ref{fig:app} shows generated samples. We only show one key generated facial gesture per example. Our project page contains videos which better reflect the quality of generations.

%% file: qualitative.tex
\begin{figure*}[t!]
\centering
\setlength{\tabcolsep}{0pt}
\begin{tabular}{p{4.7cm}|llll|p{3.9cm}|p{2.6cm}|p{3cm}}
\specialrule{1pt}{0pt}{2pt}
{\small source text} & \multicolumn{4}{l|}{\small source face sequence} & {\small true target text} & {\small text only~\cite{skipthoughts,sutskever2014sequence}} & {\small text+face}\\
\hline
%exp2
\texttt{\scriptsize we went to the hickory stick,} &
\multirow{2}{*}{\includegraphics[width=0.05\linewidth]{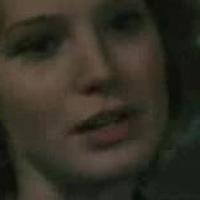}} &
\multirow{2}{*}{\includegraphics[width=0.05\linewidth]{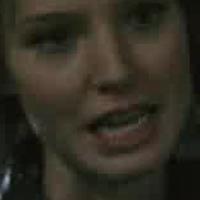}} &
\multirow{2}{*}{\includegraphics[width=0.05\linewidth]{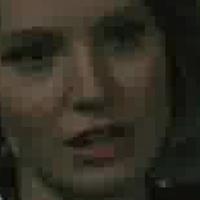}} &
\multirow{2}{*}{\includegraphics[width=0.05\linewidth]{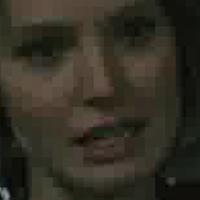}} &
\texttt{\scriptsize and then? and then i went} & 
\texttt{\scriptsize we drank a bottle} & 
\texttt{\scriptsize and then i went to}\\
\texttt{\scriptsize we had a drink, two drinks.} &
~ & ~ & ~ & ~ &
\texttt{\scriptsize home alone.} &
\texttt{\scriptsize of champagne.} & 
\texttt{\scriptsize bed.} \\
%exp3
\texttt{\scriptsize she doesn't know where he is.} &
\multirow{2}{*}{\includegraphics[width=0.05\linewidth]{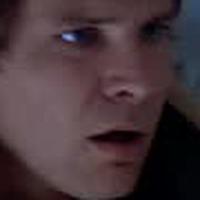}} &
\multirow{2}{*}{\includegraphics[width=0.05\linewidth]{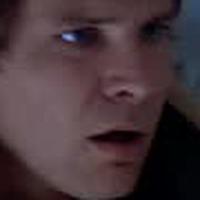}} &
\multirow{2}{*}{\includegraphics[width=0.05\linewidth]{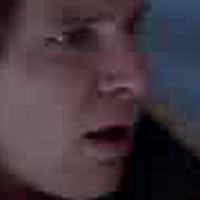}} &
\multirow{2}{*}{\includegraphics[width=0.05\linewidth]{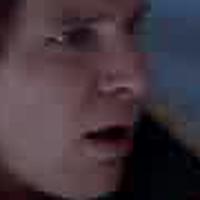}} &
\texttt{\scriptsize i don't know where he is.} & 
\texttt{\scriptsize i'm sorry.} & 
\texttt{\scriptsize i don't know where}\\
~ &
~ & ~ & ~ & ~ &
~ &
~ & 
\texttt{\scriptsize she is.} \\
%exp4
\texttt{\scriptsize and he sleeps only one hour} &
\multirow{2}{*}{\includegraphics[width=0.05\linewidth]{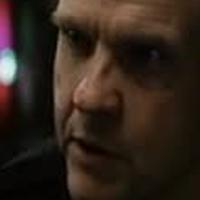}} &
\multirow{2}{*}{\includegraphics[width=0.05\linewidth]{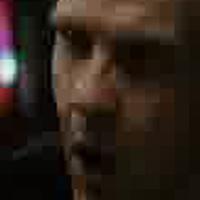}} &
\multirow{2}{*}{\includegraphics[width=0.05\linewidth]{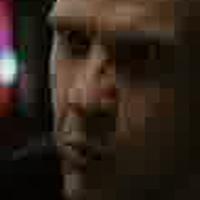}} &
\multirow{2}{*}{\includegraphics[width=0.05\linewidth]{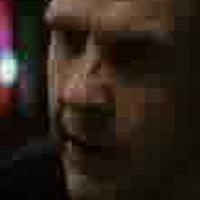}} &
\texttt{\scriptsize he's a great man.} & 
\texttt{\scriptsize he sleeps in the} & 
\texttt{\scriptsize he's a good man.}\\
\texttt{\scriptsize a night.} &
~ & ~ & ~ & ~ &
~ &
\texttt{\scriptsize same bed.} & 
~ \\
%exp5
\texttt{\scriptsize a night that marked the} &
\multirow{2}{*}{\includegraphics[width=0.05\linewidth]{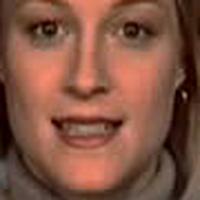}} &
\multirow{2}{*}{\includegraphics[width=0.05\linewidth]{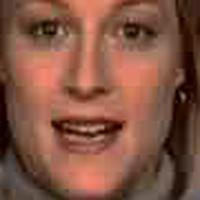}} &
\multirow{2}{*}{\includegraphics[width=0.05\linewidth]{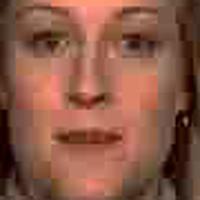}} &
\multirow{2}{*}{\includegraphics[width=0.05\linewidth]{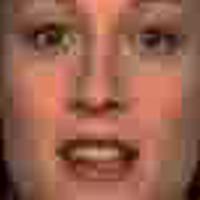}} &
\texttt{\scriptsize in world history.} & 
\texttt{\scriptsize for the future.} & 
\texttt{\scriptsize in the history of}\\
\texttt{\scriptsize opening of a new chapter.} &
~ & ~ & ~ & ~ &
~ &
~ & 
\texttt{\scriptsize the world.} \\
%exp6
\texttt{\scriptsize i hope you're not a hothead} &
\multirow{2}{*}{\includegraphics[width=0.05\linewidth]{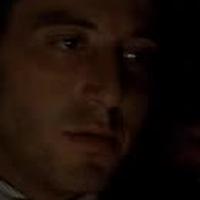}} &
\multirow{2}{*}{\includegraphics[width=0.05\linewidth]{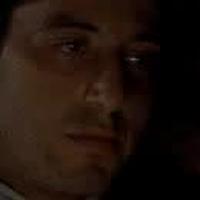}} &
\multirow{2}{*}{\includegraphics[width=0.05\linewidth]{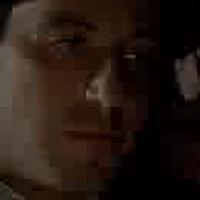}} &
\multirow{2}{*}{\includegraphics[width=0.05\linewidth]{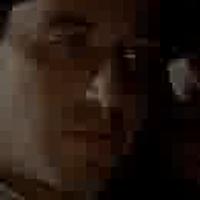}} &
\texttt{\scriptsize he's a good kid.} & 
\texttt{\scriptsize he's got a lot of} & 
\texttt{\scriptsize he's a good kid.}\\
\texttt{\scriptsize like sonny.} &
~ & ~ & ~ & ~ &
~ &
\texttt{\scriptsize something.} & 
~ \\
\hline
%exp1
\texttt{\scriptsize i guess they was worried they} &
\multirow{2}{*}{\includegraphics[width=0.05\linewidth]{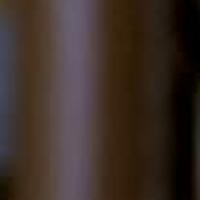}} &
\multirow{2}{*}{\includegraphics[width=0.05\linewidth]{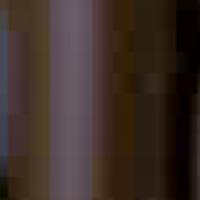}} &
\multirow{2}{*}{\includegraphics[width=0.05\linewidth]{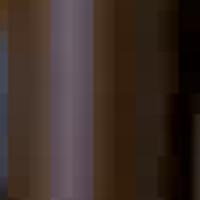}} &
\multirow{2}{*}{\includegraphics[width=0.05\linewidth]{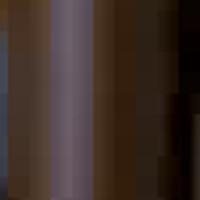}} &
\texttt{\scriptsize what's that number?} & 
\texttt{\scriptsize what's that?} & 
\texttt{\scriptsize i don't think so.}\\
\texttt{\scriptsize wouldn't find a vein in my arm.} &
~ & ~ & ~ & ~ &
~ & ~ & ~\\
%exp2
\texttt{\scriptsize oh, he's so cute.} &
\multirow{2}{*}{\includegraphics[width=0.05\linewidth]{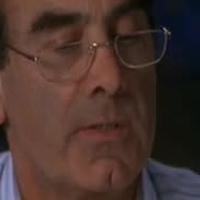}} &
\multirow{2}{*}{\includegraphics[width=0.05\linewidth]{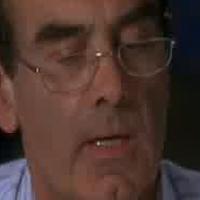}} &
\multirow{2}{*}{\includegraphics[width=0.05\linewidth]{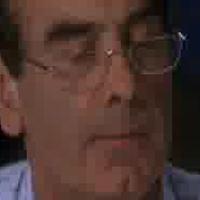}} &
\multirow{2}{*}{\includegraphics[width=0.05\linewidth]{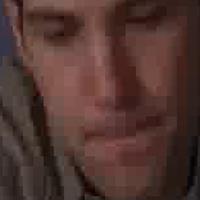}} &
\texttt{\scriptsize oh, my god.} & 
\texttt{\scriptsize oh, my god} & 
\texttt{\scriptsize he's so cute.}\\
~ &
~ & ~ & ~ & ~ &
~ &
~ & 
~ \\
%exp5
\texttt{\scriptsize can you hear me? i'm still} &
\multirow{2}{*}{\includegraphics[width=0.05\linewidth]{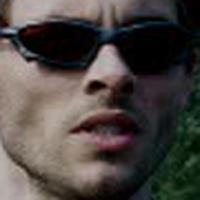}} &
\multirow{2}{*}{\includegraphics[width=0.05\linewidth]{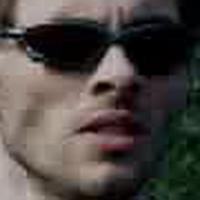}} &
\multirow{2}{*}{\includegraphics[width=0.05\linewidth]{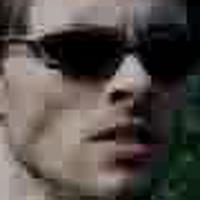}} &
\multirow{2}{*}{\includegraphics[width=0.05\linewidth]{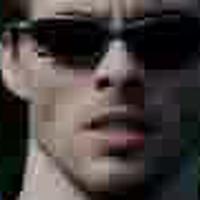}} &
\texttt{\scriptsize i'm here. scott.} & 
\texttt{\scriptsize i'm sorry.} & 
\texttt{\scriptsize what the f*** are}\\
\texttt{\scriptsize here. scott. stop.} &
~ & ~ & ~ & ~ &
\texttt{\scriptsize stop.} &
~ & 
\texttt{\scriptsize you doing here?} \\
%exp6
\texttt{\scriptsize so i don't really remember,} &
\multirow{2}{*}{\includegraphics[width=0.05\linewidth]{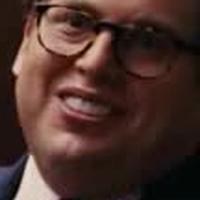}} &
\multirow{2}{*}{\includegraphics[width=0.05\linewidth]{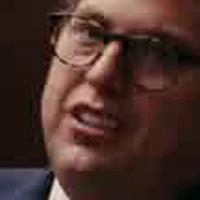}} &
\multirow{2}{*}{\includegraphics[width=0.05\linewidth]{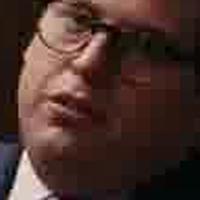}} &
\multirow{2}{*}{\includegraphics[width=0.05\linewidth]{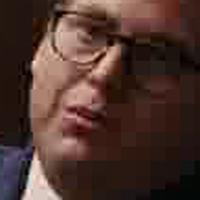}} &
\texttt{\scriptsize yeah, right. stupid.} & 
\texttt{\scriptsize yeah, yeah, yeah.} & 
\texttt{\scriptsize well, you know what?}\\
\texttt{\scriptsize yeah.} &
~ & ~ & ~ & ~ &
~ &
~ & 
\texttt{\scriptsize i'm sorry.} \\
%exp7
\texttt{\scriptsize i can't feel my legs.} &
\multirow{2}{*}{\includegraphics[width=0.05\linewidth]{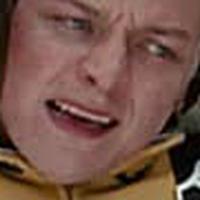}} &
\multirow{2}{*}{\includegraphics[width=0.05\linewidth]{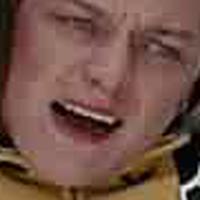}} &
\multirow{2}{*}{\includegraphics[width=0.05\linewidth]{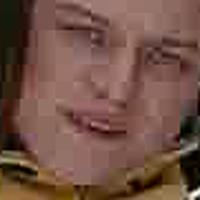}} &
\multirow{2}{*}{\includegraphics[width=0.05\linewidth]{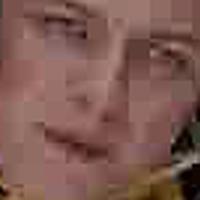}} &
\texttt{\scriptsize i can't feel my legs.} & 
\texttt{\scriptsize and i can't} & 
\texttt{\scriptsize it's too much.}\\
~ &
~ & ~ & ~ & ~ &
~ &
\texttt{\scriptsize breathe.} & 
~ \\
\specialrule{1pt}{2pt}{8pt}
\end{tabular}
\vspace{-6mm}
\caption{\small Success and failure cases of using face along with text. Top five rows show successful examples where adding facial gesture information produces sentences closer to the ground truth. Bottom five rows show failure modes, including face detection failure in the sixth row, and detecting another face that does not belong to the speaker in the seventh row.}
\vspace{-0mm}
\label{fig:quali}
\end{figure*}

%% file: hankfig.tex
\begin{figure*}
\centering
\setlength{\tabcolsep}{1pt}
\setlength\extrarowheight{-3pt}
\begin{tabular}{lp{5.4cm}lp{5.4cm}}
\hline
\multirow{8}{*}{\includegraphics[width=0.144\linewidth]{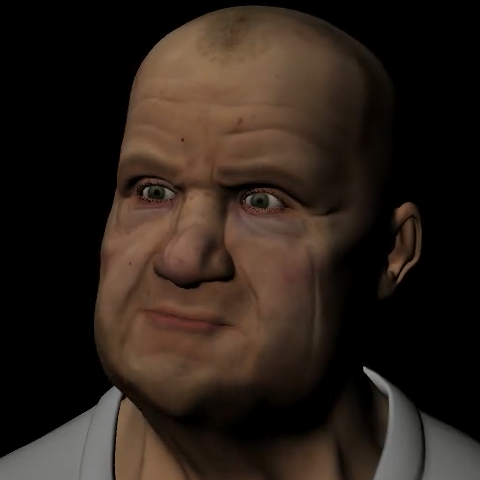}}$\ $ &
\texttt{\scriptsize Query: you are my creator.} &
\multirow{8}{*}{\includegraphics[width=0.144\linewidth]{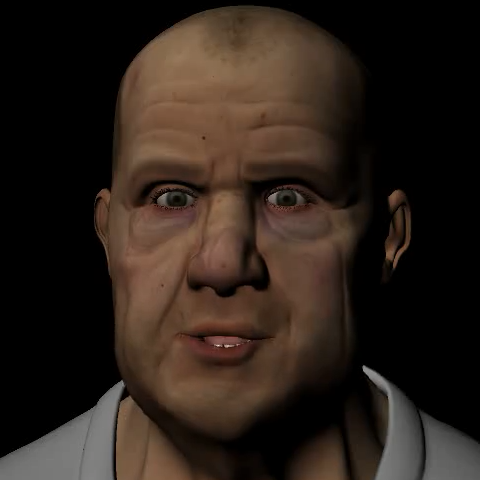}}$\ $  &
\texttt{\scriptsize Query: can't wait to hear them.} \\
~ &
\texttt{\scriptsize} &
~ &
\texttt{\scriptsize first things first.} \\
\cline{2-2}\cline{4-4}
~ &
\texttt{\scriptsize A1: you are my savior.} &
~ &
\texttt{\scriptsize A1: what's your name?} \\
~ &
\texttt{\scriptsize} &
~ &
\texttt{\scriptsize} \\
\cline{2-2}\cline{4-4}
~ &
\texttt{\scriptsize A2: i am the something.} &
~ &
\texttt{\scriptsize A2: i can't do that.} \\
~ &
\texttt{\scriptsize} &
~ &
\texttt{\scriptsize} \\
\cline{2-2}\cline{4-4}
~ &
\texttt{\scriptsize A3: and i will show you my} &
~ &
\texttt{\scriptsize A3: well, then, i can see} \\
~ &
\texttt{\scriptsize power.} &
~ &
\texttt{\scriptsize you're a little confused.} \\
\hline\hline
\multirow{8}{*}{\includegraphics[width=0.144\linewidth]{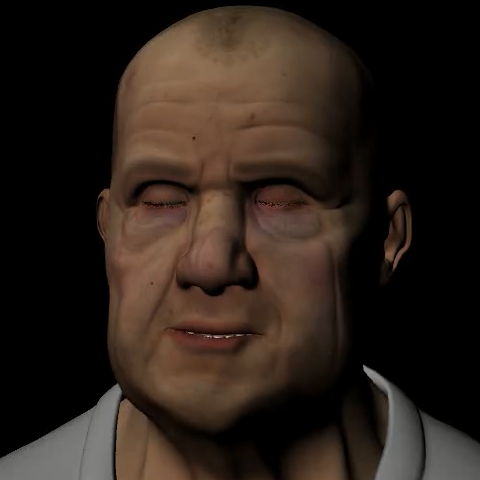}}$\ $  &
\texttt{\scriptsize Q: i'd suffer the} &
\multirow{8}{*}{\includegraphics[width=0.144\linewidth]{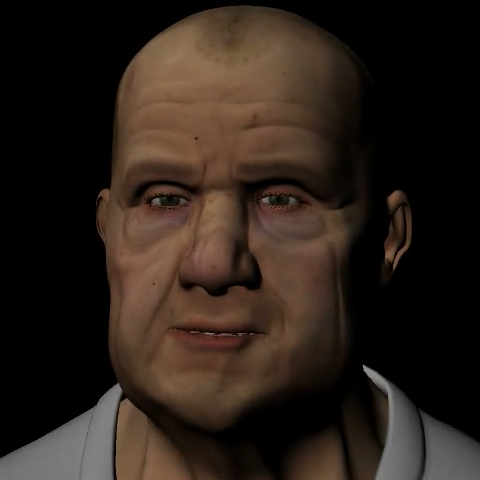}}$\ $  &
\texttt{\scriptsize Q: with your head blown off. do} \\
~ &
\texttt{\scriptsize consequence.} &
~ &
\texttt{\scriptsize you understand what i'm saying?} \\
\cline{2-2}\cline{4-4}
~ &
\texttt{\scriptsize A1: i'm sorry.} &
~ &
\texttt{\scriptsize A1: yes, sir.} \\
~ &
\texttt{\scriptsize} &
~ &
\texttt{\scriptsize} \\
\cline{2-2}\cline{4-4}
~ &
\texttt{\scriptsize A2: i can't.} &
~ &
\texttt{\scriptsize A2: you know, i don't know} \\
~ &
\texttt{\scriptsize} &
~ &
\texttt{\scriptsize what i'm doing.} \\
\cline{2-2}\cline{4-4}
~ &
\texttt{\scriptsize A3: i would rather die than} &
~ &
\texttt{\scriptsize A3: we're gonna shred the} \\
~ &
\texttt{\scriptsize be with you.} &
~ &
\texttt{\scriptsize evidence first.} \\
\hline\hline
\multirow{8}{*}{\includegraphics[width=0.144\linewidth]{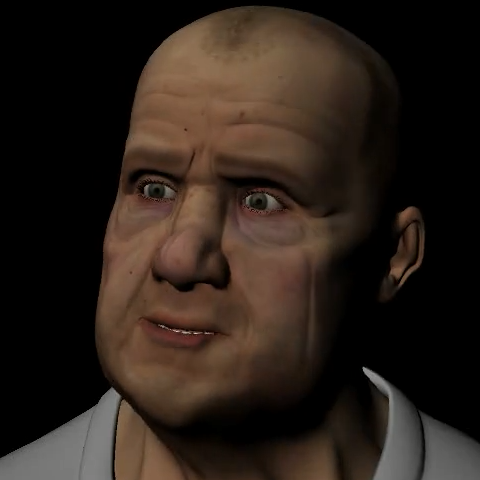}}$\ $  &
\texttt{\scriptsize Q: he was supposed to kick back} &
\multirow{8}{*}{\includegraphics[width=0.144\linewidth]{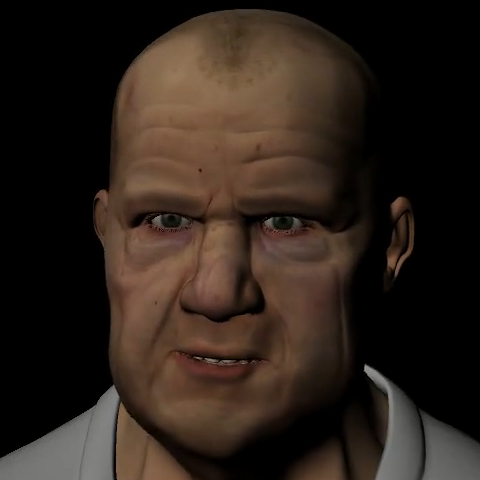}}$\ $  &
\texttt{\scriptsize Q: you're not getting boy with} \\
~ &
\texttt{\scriptsize for or five mil, was it?} &
~ &
\texttt{\scriptsize apple, you goddamn little fruit!} \\
\cline{2-2}\cline{4-4}
~ &
\texttt{\scriptsize A1: yeah.} &
~ &
\texttt{\scriptsize A1: i'm sorry.} \\
~ &
\texttt{\scriptsize} &
~ &
\texttt{\scriptsize} \\
\cline{2-2}\cline{4-4}
~ &
\texttt{\scriptsize A2: i'm saying he's a good guy} &
~ &
\texttt{\scriptsize A2: you're not gonna be a} \\
~ &
~ &
~ &
\texttt{\scriptsize f***ing retard, man.} \\
\cline{2-2}\cline{4-4}
~ &
\texttt{\scriptsize A3: do you know how much i} &
~ &
\texttt{\scriptsize A3: you said you, you} \\
~ &
\texttt{\scriptsize love you?} &
~ &
\texttt{\scriptsize something me.} \\
\hline
\end{tabular}
\vspace{-1mm}
\caption{\small NeuralHank examples. \texttt{Q} is the query text. \texttt{A1}, \texttt{A2}, and \texttt{A3} are generated by \textit{noMicro-noGAN}, \text{Micro-noGAN}, and \textit{Ours}, respectively. We also show one animation frame generated by our method. First two rows show that our GAN-based model generates more diverse and interesting responses. Last row shows failure cases where our method generates confusing responses. Please refer to our project page for animation videos, more examples, and chatting with Hank live through a webcam.}
\vspace{-4mm}
\label{fig:app}
\end{figure*}

%% file: conc.tex
\section{Conclusion}
\label{sec:conc}

We proposed a face-to-face neural conversation model, an encoder-decoder neural architecture trained with RL and GAN. Our approach used both textual and facial information to generate more appropriate responses for the conversation. We trained our model by exploiting rich video data in form of movies. We evaluated our model through a mind-reading test as well as a virtual chatting avatar. In the future, we aim to learn body controllers as well, model the personalities of the conversation participants, as well as capture more high-level semantics of the situation~\cite{moviegraphs}.